# ApproXAI: Energy-Efficient Hardware Acceleration of Explainable AI using Approximate Computing

Ayesha Siddique, Khurram Khalil, Khaza Anuarul Hoque
*Department of Electrical Engineering and Computer Science*
*University of Missouri-Columbia, USA*
{ayesha.siddique, khurram.khalil, hoquek}@missouri.edu

*Abstract*—Explainable artificial intelligence (XAI) enhances AI system transparency by framing interpretability as an optimization problem. However, this approach often necessitates numerous iterations of computationally intensive operations, limiting its applicability in real-time scenarios. While recent research has focused on XAI hardware acceleration on FPGAs and TPU, these methods do not fully address energy efficiency in real-time settings. To address this limitation, we propose XAIedge, a novel framework that leverages approximate computing techniques into XAI algorithms, including integrated gradients, model distillation, and Shapley analysis. XAIedge translates these algorithms into approximate matrix computations and exploits the synergy between convolution, Fourier transform, and approximate computing paradigms. This approach enables efficient hardware acceleration on TPU-based edge devices, facilitating faster real-time outcome interpretations. Our comprehensive evaluation demonstrates that XAIedge achieves a $2\times$ improvement in energy efficiency compared to existing accurate XAI hardware acceleration techniques while maintaining comparable accuracy. These results highlight the potential of XAIedge to significantly advance the deployment of explainable AI in energy-constrained real-time applications.

*Index Terms*—Approximate computing, explainable artificial intelligence, deep neural networks, hardware accelerator, multi-pod systolic array accelerators.

## I. INTRODUCTION

Explainable AI (XAI) is a promising paradigm for improving the interpretability of AI-driven decisions [1]. It provides detailed insights into the reasoning behind the predictions and actions of black-box models, which is crucial for building user trust in safety-critical applications. For example, in malware detection, understanding the rationale behind labeling software as malicious or benign is as important as the classification itself. In such applications, XAI can help identify threats and elucidate their nature and precise location within the software [2]. This capability of XAI significantly improves the efficacy and accountability of AI applications, enabling more targeted and effective responses [3], [2]. XAI algorithms, such as integrated gradients, Shapley values, etc., are typically implemented on FPGA, GPU, and TPU-based hardware accelerators [4], [5], [6]. The recent advent of Google Egde TPU necessitates the adoption of energy-efficient XAI algorithms. However, XAI algorithms consider the explanation process an additional step alongside the prediction, which can incur computational costs, energy consumption, and time overhead. This makes XAI less suitable for energy-constrained real-time applications where both speed and energy efficiency are critical. For example, lagging interpretability in multimedia and gaming applications can degrade the quality of service. In contrast, an energy outage in safety-critical real-time systems can endanger human lives. *This is a significant bottleneck that can restrict the widespread adoption of XAI, underscoring the need for more universally adaptable explanation techniques on energy-constrained hardware accelerators.*

### A. Current State-of-the-art and Limitations

Recent advancements in AI hardware have contributed significantly to the acceleration of XAI algorithms. Mitchell et al. proposed GPUTreeShap, an adaptation of the TreeShap algorithm optimized for massively parallel processing of XAI algorithms on GPUs [4]. Zhou et al. introduced a dedicated hardware architecture tailored to enhance XAI performance in graph-convolutional networks using field-programmable gate arrays (FPGAs) [5]. Bhat et al. proposed a feature attribution algorithm based on gradient backpropagation designed for XAI and implemented the entire framework in FPGA to achieve execution on the edge [7]. In another work, the author [8] proposed a novel non-uniform interpolation scheme to compute the attribution scores in integrated gradients on GPUs. Recently, Pan et al. leveraged TPUs to further accelerate the explainability process by translating it into Fast Fourier Transform (FFT) and polynomial interpolation methods [6]. However, these state-of-the-art methods for XAI hardware acceleration have the following limitations:

1) The hardware acceleration of XAI techniques necessitates specific FPGA customization for each application, which constrains their broad applicability in various scenarios.
2) Translating XAI techniques to mathematical models, such as FFT and polynomial interpolation, presents significant computational challenges due to the inherently energy-intensive nature of these models, which involve matrix multiplications.
3) Furthermore, target hardware accelerators like Google Edge TPU have yet to be considered. Google Edge TPU is more energy-constrained than Google TPU, making the energy-efficient hardware acceleration of XAI algorithms even more challenging.

These observations raise a crucial research challenge: "*How can we enhance the speed and energy efficiency of the XAI algorithms to make them suitable for real-time applications on energy-constrained edge devices?*"

## B. Novel Contributions

To address the above research challenge, we leverage approximate computing in XAI algorithms for their energy-efficient parallel hardware acceleration on TPU-based edge devices (XAIedge). Approximate computing seeks to improve energy efficiency by enabling controlled imprecision in computational tasks. More specifically, our novel contributions are as follows.

1) A novel **design framework** (XAIedge) that leverages energy-aware fine-tuned approximate computing into the XAI algorithms, namely, model distillation, integrated gradients, and Shapley values with the parallel hardware acceleration of matrix operations on hardware accelerators.
2) A **comprehensive performance analysis** of the proposed framework for energy-aware approximate XAI hardware acceleration in terms of accuracy-energy tradeoffs on CPU, GPU, and TPU-based hardware accelerators. This also includes a comparison with the state-of-the-art.

To demonstrate the effectiveness of our proposed framework XAIedge, we used state-of-the-art classifiers (VGG19, ResNet50, MobileNetV2) on diverse datasets (CIFAR10 [9], MIRAI [10], ImageNet). Our results show that by implementing approximate FFT and approximate matrix multiplications [11], XAIedge achieves a $2\times$ improvement in energy efficiency compared to existing techniques while maintaining comparable accuracy. Note, approximate computing has been conventionally used for energy-efficient hardware acceleration of deep neural networks [12], [13], [14]. However, to the best of our knowledge, this is the first successful attempt to leverage approximate computing for energy-efficient XAI hardware acceleration, opening new avenues for XAI in resource-constrained edge environments.

## II. BACKGROUND

This section provides the preliminary information related to model distillation, integrated gradients, and Shapley values.

### A. Model Distillation

Model distillation in the context of XAI is a technique that constructs a simpler distilled model $\theta^*$ to approximate the complex relationship between input features $x$ and outputs $y$ of a complex machine learning model $\theta$. This is achieved by identifying the optimal parameters $\mathcal{P}$ for the target model $\theta$ by solving an optimization problem as given below.

$$\mathcal{P} = \arg\min_{\mathcal{P}} \|\theta^*(x) - y\| \quad (1)$$

As an example, let's assume the model $\theta^*$ is distilled into a decision tree. Decision trees offer a clear, rule-based structure that makes the logic behind each classification decision easily traceable from root to leaf. By examining the paths taken for different classifications, we can identify which pixel values and patterns are most influential in leading to specific outcomes. This method provides a clear visualization of how input features drive the model toward the final outcome.

### B. Integrated Gradients

Integrated gradients is a method that explains ML model predictions by tracking how the model's output changes from a baseline input (a reference input for which the prediction is known and typically represents the absence of features) to the actual input. This is done by interpolating between the baseline and actual input and then accumulating the gradients of the model's output with respect to the input features along this path. The integrated gradients attribution for a feature $i$ of an input record $x$ relative to a baseline record $x'$ is given by the following equation:

$$IG_i(x) = (x_i - x_i') \times \int_{\alpha=0}^{1} \frac{\partial f(x' + \alpha \times (x - x'))}{\partial x_i} d\alpha \quad (2)$$

where $IG_i(x)$ is the attribution of feature $i$ for the input $x$. $x_i$ and $x_i'$ are the values of feature $i$ in the actual and baseline inputs, respectively. $f$ denotes the model's prediction function. $\frac{\partial f(x)}{\partial x_i}$ is the gradient of $f$ with respect to the feature $i$. $\alpha$ scales the interpolation between the baseline $x'$ and the input $x$.

### C. Shapley Analysis

Shapley analysis is derived from cooperative game theory. The computation of Shapley values for machine learning tasks involves treating features as 'players' in a game. The attribution of each feature is determined by the weighted average of the marginal contribution across all possible feature subsets. Mathematically, the Shapley value for the $i$-th feature is calculated as:

$$\phi_i = \sum_{S \subseteq M \setminus \{i\}} \frac{|S|!(|M| - |S| - 1)!}{|M|!} [f_x(S \cup \{i\}) - f_x(S)] \quad (3)$$

where $M$ and $S$ represent a total number of features and a subset of M that excludes the $i$-th feature, respectively. $f_x(\cdot)$ denotes the model prediction function for a subset of features. Let's assume a logistic regression model designed to predict the likelihood of a patient having a heart disease based on three key features: cholesterol level, blood pressure, and smoking status. Considering all 3! = 6 permutations of features, Shapley values can be calculated by averaging the contribution of each feature to the model's prediction. Specifically, starting with a baseline prediction that reflects the average outcome without feature specifics, the model sequentially integrates each feature, observing changes through cross-entropy loss to assess marginal contributions. This process highlights the relative importance of each feature, identifying cholesterol as the most significant predictor, followed by blood pressure and smoking status. Thus, Shapley analysis offers a precise and equitable assessment of feature impact, enhancing understanding of the model's decision making in heart disease prediction.

## III. PROPOSED XAIEDGE FRAMEWORK FOR APPROXIMATE XAI HARDWARE ACCELERATION

Our proposed framework, XAIedge, focuses on fine-tuned approximate hardware acceleration of key XAI techniques: model distillation, integrated gradients, and Shapley analysis.

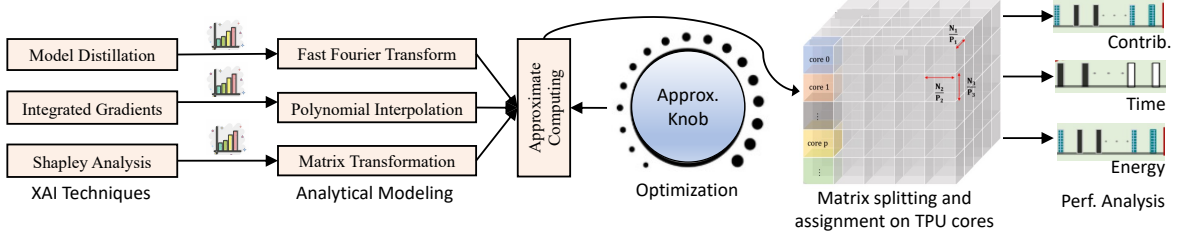

Fig. 1: Our Proposed Framework XAIedge for Approximate XAI Hardware Acceleration

**Algorithm 1:** Approximate Model Distillation

**Inputs**: Input data: $x$; Perturbed input: $x'$; Model's output: $y$
**Outputs:** Contribution factor: $\mathcal{F}$
1: $\mathcal{D}$ = op_accel ($x$, $FFT$, @minlp)
2: $\mathcal{F}_y$ = op_accel ($\mathcal{D}$, fft, @minlp)
3: $\mathcal{F}_x$ = fft_1d ($y$)
4: $d_k = \mathcal{F}_y / \mathcal{F}_x$
5: $\theta_k$ = op_accel($d_k$, $IFFT$, @minlp)
6: $z = x' * \theta_k$
7: $\mathcal{C} = y - z$
8: **return** $\mathcal{C}$

As shown in Fig. 1, the design flow of our porposed framework consists of four key computational steps: (i) First, XAI algorithms are translated into fundamental matrix operations utilizing FFT and polynomial interpolation. (ii) Next, approximate multipliers are introduced as a substitute for accurate multipliers in the computations. (iii) The approximation level of approximate multiplier is determined using Mixed Integer Nonlinear Programming (MINLP) [15]. (iv) The matrix operations are split into multiple subtasks and executed in parallel on separate TPU cores without requiring any data exchange between cores. (iv) Finally, trade-offs between accuracy and energy efficiency are analyzed on the basis of user-defined accuracy and energy criteria. The following sections discuss the approximate hardware acceleration of approximate model distillation, integrated gradients, and Shapley analysis.

**A. Approximate Model Distillation:** Model distillation is a technique that creates a simpler distilled model $K^*$ to approximate the complex relationship between input features $X$ and outputs $Y$ of a sophisticated machine learning model $K$, a multi-layer convolutional neural network in our case. To achieve energy-efficient hardware acceleration of model distillation, we compute the distilled model by leveraging approximate computing-based FFT (detailed in Section III.D) and harnessing the parallel processing capabilities of TPU. Mathematically, the relationship between input $X$ and output $Y$ of a distilled model can be represented as $X * K = Y$, where $*$ denotes the convolution operator. The distilled model can be computed as $K = \mathcal{F}_a^{-1}(\mathcal{F}_a(Y)/\mathcal{F}_a(X))$, where $\mathcal{F}_a$ and $\mathcal{F}_a^{-1}$ represent approximate Fourier and approximate inverse Fourier transform. The distilled model $K$ is inherently explainable and enables the users to discern the decision rules or identify the important input features influencing the model's outputs $Y$. Hiding important input features may cause considerable changes to the model output. Considering this, we remove a target feature $x_i$ from the original input $X = [x_1, x_2, ..., x_{i-1}, x_i, x_{i+1}, ..., x_d]$ to generate a perturbed input as $X' = [x_1, x_2, ..., x_{i-1}, 0, x_{i+1}, ..., x_d]$. Then, by calculating the difference between the model's output and perturbed input, we find the contribution factor $C$ of $x_i$ as $\mathcal{C}(x_i) \triangleq Y - X' * K$.

**Algorithm 2:** Acceleration of Integrated Gradients

**Inputs**: Input tensor: $x$; Baseline input: $x'$; Steps: $n$; No. of TPU cores: $p$; Approx. multiplier: $M$; Machine learning model: $\theta$
**Outputs:** Resulting attributes: $\mathcal{T}$
1: $\delta = x - x'$
   // Find difference between input and baseline
2: $x^* = x' + \delta \cdot \left(\frac{\text{float}(i)}{n-1}\right) \mid i = (0, 1, \ldots, n-1)$
   // Generate interpolated inputs
3: $y = [\theta(p_x) \mid p_x \in x^*]$
   // Calculate predictions for the interpolated inputs
4: $\mathcal{V}$ = ComputeVendermode($n$)
   // Create Vandermonde matrix for the interpolation points
5: $\mathcal{F}$ = ComputePol($\mathcal{V}$, $y$, $p$, $M$, @op_accel, @minlp)
   // Compute coefficients using approximate multiplier in distributed matrix mutliplication over multiple TPU cores
6: **for** $k$ in range (1, $t$+1) **do**
7:    $\alpha_{k-1}$ = ($k$-1) /$t$
8:    $\alpha_k$ = $k/t$
9:    $\mathcal{G}$ = TrapezoidalIntegration ($\mathcal{F}$, $\alpha_{k-1}$, $\alpha_k$)
      // Approximate the IGs over each sub-interval
10: **end for**
11: $\mathcal{T} = \mathcal{G} \cdot \delta$ // Compute the final attribution
12: **return** $\mathcal{T}$

Algorithm 1 delineates the steps involved in the hardware acceleration of approximate model distillation on TPU. This algorithm takes input data $x$, perturbed input data $x'$, and the model's output $y$ as inputs. First, a 2-stage approximate 2D FFT is applied to input data. The 2-stage approximate 2D FFT refers to an accelerated 1D-FFT function, @op_accel function in Algorithm 5, that is applied in both dimensions of the input data (Line 1-2). This function exploits approximate computing in FFT and inverse FFT, and leverages the parallel processing capabilities of TPU-based edge devices. The approximation levels for the approximate FFT are determined using the MINLP solver. Next, a 1-dimensional FFT is applied to the target model's output vector (Line 3). Then, the distilled model

**Algorithm 3:** Approximate Shapley Value Analysis

**Inputs :** *Input tensor: $x$; baseline input: $x'$;*
*Number of cores: $p$; Model: $\theta$*
**Outputs:** Shapley values: $values$

1: $values = [\ ]$
2: $n = x.\text{shape}[1]$
3: **for** $i \in \{1, \ldots, n\}$ **do**
4:   **for** $S \in \mathcal{P}(\{1, \ldots, n\})$ **do**
5:     **if** $i \notin S$ **then**
6:       $S^{+i} = S \cup \{i\}$
7:       $S^{-i} = S$
8:       $x_{new} = x'$
9:       $x_{new}[:, S^{+i}] = x[:, S^{+i}]$
10:      $x^*_{new} = x'$
11:      $x^*_{new}[:, S^{-i}] = x[:, S^{-i}]$
12:      $\mathring{f}_{S+i} = \text{predict}(\theta, x_{new})$
13:      $\mathring{f}_{S-i} = \text{predict}(\theta, x^*_{new})$
14:      $cb = \mathring{f}_{S+i} - \mathring{f}_{S-i}$
15:      $temp = weight(S, i, n) * cb, p, @op\_accel)$
16:      $values \mathrel{+}= temp$
       // Update the Shapley value for $i^{th}$ feature
17:     **end if**
18:   **end for**
19: **end for**
20: **return** $values$

---

**Algorithm 4:** Approximate FFT (**@ax_fft**)

**Inputs :** Input data: $x$; Multiplier: $M$;
List of approximation levels: $levels$;
PSNR threshold: $p_t$; Energy threshold: $e_t$
**Outputs:** fft result: $\mathcal{R}$

1: $\mathcal{R} = \mathbf{0}_n \mid (n = \text{len}(x))$
2: **for each** $t$ in range ($\log_2(n)$) **do**
3:   $m = 2^t$
4:   $S = n\ /\ m$
5:   **for each** $s$ in range($S$) **do**
6:     **for each** $i$ in range($s \cdot m$, $(s \cdot m) + m/2$) **do**
7:       $r = i + m/2$
        // Apply butterfly operation
8:       $l = levels[t]$
        // Level for the current stage
9:       $\mathcal{B} = \text{approx\_multiply}(x(i), x(r), l, M,$
        @op_accel, @minlp)
        // Approx. multiplication for butterfly output
10:      $x(i) = x(i) + \mathcal{B}$
        // Update signal array with the butterfly operation
11:      $x(r) = x(i) - 2 * \mathcal{B}$
12:     **end for**
13:   **end for**
14: **end for**
15: $\mathcal{R} = x$
16: **return** $\mathcal{R}$

---

is computed (Lines 4-5), followed by the calculation of the contribution factor for the perturbed input (Lines 6-7). Finally, the algorithm returns the contribution factor as an output.

**B. Approximate Integrated Gradients:** Integrated Gradients is a method that explains ML model predictions by tracking how the model's output changes from a baseline input (a reference input for which the prediction is known and typically represents the absence of features) to the actual input. The integrated gradients method faces challenges in analytically solving the integral of complex output functions in neural networks. We address this by employing numerical integration with polynomial interpolation, using the Vandermonde matrix for interpolation. To mitigate the computational and energy costs associated with solving the Vandermonde system, we distribute matrix operations across multiple TPU cores, applying an approximate multiplier in each core for reduced energy consumption. Our approach partitions the integration interval, applies the trapezoidal rule to each sub-interval, and aggregates the results. Algorithm 2 outlines the steps for hardware acceleration of approximate integrated gradients on TPU-based edge devices. The inputs to this algorithm are: an input tensor, a baseline input, number of steps (or intervals) that partition the path from the baseline to the input, number of TPU cores, approximate multiplier, and the target machine learning model. The algorithm begins by calculating the difference between the input tensor and the baseline input (Line 1), followed by generating the interpolated points (Line 2). Predictions for these interpolated inputs are then generated using the target model (Line 3). Subsequently, the Vandermonde matrix is computed based on the specified number of interpolation points (Line 4). Next, the polynomial approximation coefficients are computed in parallel by distributing the matrix multiplications across all TPU cores. Specifically, each matrix of size $a \times b$ is divided into smaller sub-matrices using @op_accel function to enable parallel computation across $p$ cores, rather than using a single core (Line 5). The computation of polynomial coefficients involves energy-intensive multiplication of the inverse Vandermonde matrix with the model's output vector for the interpolated inputs. To optimize energy efficiency, a MILP solver is employed alongside approximate multipliers. After this, the integrated gradients are computed over each sub-interval (Lines 7-10), where both the normalized coefficients for the current and previous steps are used alongside the approximate polynomial coefficients. For the integrated gradients, the trapezoidal integration is utilized by averaging the gradients at consecutive interpolation points to improve the accuracy of the integral approximation. Finally, the approximate gradients are multiplied by the difference between the input tensor and the baseline input to compute the final attributions (Line 11), which form the final output.

**C. Approximate Shapley Analysis:** Shapley analysis quantifies the contribution of each feature to a model's prediction by distributing the output among the features based on cooperative game theory. Algorithm 3 outlines the hardware acceleration of approximate Shapley analysis on TPU-based devices. This algorithm takes an input tensor $x$, a baseline input $x'$, a number of TPU cores and a trained model. The algorithm begins with initializing an empty list to store the Shapley values (Line 1) and determining the number of features in the input tensor (Line 2). For each feature index $i$, the algorithm considers all possible subsets that correspond to combinations of feature indices excluding $i$ (lines 3 and 4). For each such subset $S$, two index sets are defined: $S^{-i}$ representing the original subset without feature index $i$, and $S^{+i}$, representing the same subset with feature index $i$ (Lines 6 and 7). Next, two variants of the input tensors are constructed: $x_{new}$ which takes feature values from $x$ at indices in $S^{+i}$, and $x*_{new}$ which takes feature values from $x$ at indices in $S^{-i}$. In both cases, the remaining features are filled with the values from the baseline input $x'$ (Lines 8-11). The marginal contribution of the feature is then computed as the difference in model predictions on these two tensors (Lines 12-14). This contribution is weighted using the Shapley value coefficient, which is computed using factorial-based weights (Line 15). These computations are distributed across multiple TPU cores using @$op\_accel$ to improve efficiency. Finally, the Shapley values for each feature are aggregated (Line 16) and returned as the output.

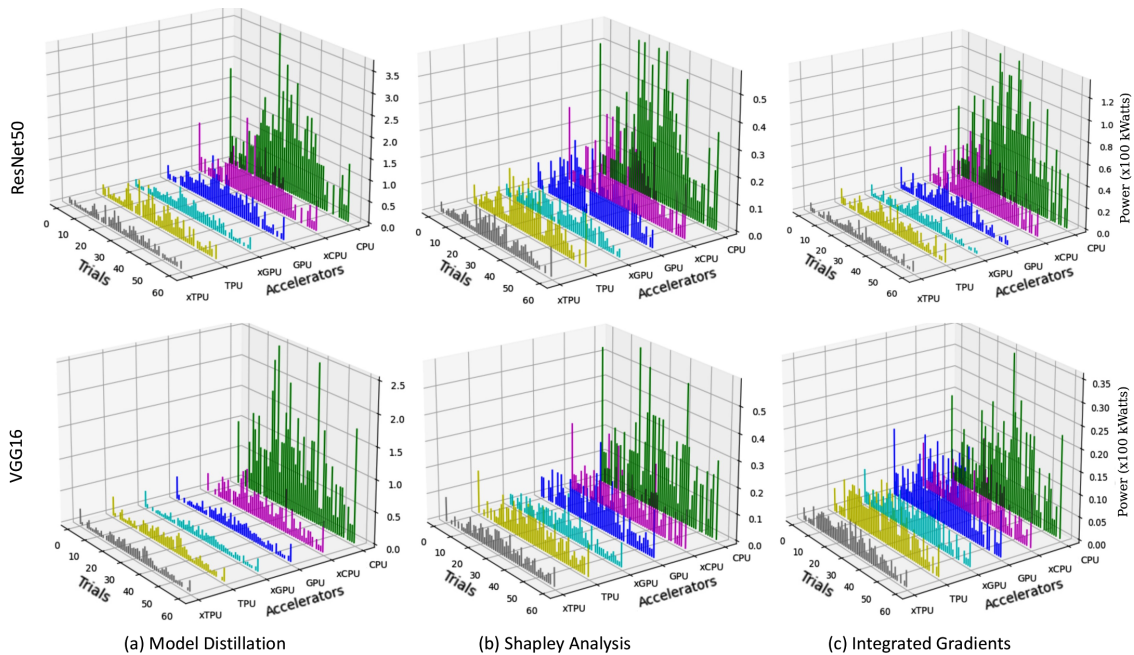

Fig. 2: Power consumption of approx. hardware acceleration of XAI algorithms i.e., model distillation, Shapley analysis, and integrated gradients on CPU, GPU, and TPU with accurate multipliers and on xCPU, xGPU, and xTPU with approx. multipliers.

**D. Approximate Fast Fourier Transform:** The approximate hardware acceleration of the model distillation method is based on approximate FFT that utilizes the tunable approximate multiplier to update the approximation level according to the desired peak signal-to-noise ratio (PSNR). The approximation level $n_{stage}$ of approximate FFT is determined using the MINLP [15]. The MINLP finds the optimal $n_{stage}$ that satisfy the following objective function:

$$\text{Minimize} \sum n_{stage} \quad (4)$$

with subject to $p_c > p_t$, $e_c > e_t$, $n_{stage} \in [0, 11]$,

$$P(x | x \in S_r \cap S_e) \geq P_t,$$

where $p_c$ and $e_c$ are the calculated PSNR and energy for the approximate XAI techniques, $p_t$ and $e_t$ are the PSNR and energy thresholds, and $P_t$ is the probability constraint. $S_r$ and $S_e$ represent sets of PSNR and energy consumption values obtained for $s$ input sequences, respectively.

Algorithm 4 outlines the steps for approximate 1D FFT. This algorithm takes input data $x$, multiplier, list of approximation levels for each stage, PSNR threshold, and energy threshold. It begins with initializing an array $\mathcal{R}$ to zeros, with the same length as the input data $x$ (Line 1). The FFT is computed in $log_2(n)$ stages, where $n$ is the length of the input data (Line 2). In each stage $t$, the input data is divided into equal-sized segments of length $2^t$, and the number of such segments is calculated as $n/m$ (lines 3 and 4). For each segment, the algorithm iterates over the first half of its indices $i$ and computes the corresponding butterfly index $r$ (Line 5-7). These index pairs define the positions of the elements involved in the butterfly operation. Next, the approximation level for the current stage is retrieved from $levels$ (Line 8). In each butterfly operation, the accurate multiplication is replaced by the approximate multiplier, and the approximation levels are optimized using the MINLP to satisfy PSNR and energy constraints (Lines 9). The result of this multiplication is used to increment the value at index $i$ in $x$ (Line 10). This updated value is then used to compute a scaled difference, which replaces the value in index $r$ (Line 11). This process is repeated across all index pairs, segments, and stages. After completing all updates, the modified input data is assigned to $\mathcal{R}$ (Line 15). Finally, the algorithm returns $\mathcal{R}$ as the output, representing the approximate FFT of the input data.

For hardware acceleration, we implement the $op\_accel$ function (Algorithm 5) to implement the operations such as

**Algorithm 5:** Accelerated FFT/ IFFT/ Poly./ Factorial (@**op_accel**)

**Inputs** : *Input data: $x$; Number of cores: $p$;*
*Desired operation: $\mathcal{O}$ = {fft or ifft or poly};*
*Approximate multiplier: $M$*

**Outputs:** *Resulting matrix: $y$*

1: $x_{split}$ = split ($x$, $p$, axis=1)
   // Apply desired operation on each core
2: $\mathcal{L}_{core}$ = [ ]
   // Initialize a list to store results from each core
3: **for each** i in $p$ **do**
4:    split_p = $x_{split}[p]$
   Extract data to be processed on $p^{th}$ core
5:    **if** $\mathcal{O}$ = fft **then**
6:      $\mathcal{D}$ = fft($x_{split}$, $M$), @milp))
7:    **else if** $\mathcal{O}$ = ifft **then**
8:      $\mathcal{D}$ = ifft($x_{split}$, $M$), @milp))
9:    **else if** $\mathcal{O}$ = poly **then**
10:     $\mathcal{D}$ = poly($x_{split}$, $M$), @milp))
11:   **end if**
12:   $\mathcal{L}_{core}$ = $\mathcal{L}_{core} \cup \mathcal{D}$
    // Append result from current core
13: **end for**
14: $y$ = concatenate($\mathcal{L}_{core}$, axis=0)
    // Merge data from all cores
15: $y'$ = transpose($y$)
    // Transpose the final result
16: **return** $y'$

1-D FFT, inverse FFT, or polynomial interpolation in parallel across multiple TPU cores. This function first splits an $M \times N$ input matrix into sub-matrices along the row dimension, such that each available TPU core processes one sub-matrix (Line 1). Next, an empty list is initialized to collect outputs from all cores (Line 2). Each core processes the assigned sub-matrix according to the specified operation (Lines 5-10). All operations utilize a fine-tuned floating-point (FP) approximate multiplier [11] implemented in Intel's bfloat16 format. The approximation level of the approximate multiplier is configured by the MINLP solver. Once all operations are complete, the results from all cores are gathered and merged into a new $M \times N$ matrix (Line 14). Finally, the resulting matrix is transposed and returned as the final output (Lines 15 and 16). This algorithm is integrated into model distillation, integrated gradients and Shapley analysis to support parallel matrix multiplication using the approximate multiplier on multiple TPU cores.

## IV. RESULTS AND DISCUSSIONS

This section comprehensively analyzes our proposed XAIedge detailing the experimental setup and key findings.

### A. Datasets and Architectures

The performance of our proposed framework XAIedge is evaluated using three well-known classifiers: VGG16 (CIFAR100), MobileNetV2 (ImageNet), and ResNet50 (NSS MIRAI malware detection). Table I presents the classification accuracy of each model considered for approximate XAI hardware acceleration on CPU, GPU, and TPU-based hardware accelerators. The reported accuracy values are aligned with the state-of-the-art. Our experimental setup uses Python 3.10.12 and Tensorflow 2.15.0 on Google Colab, featuring TPU v2, NVIDIA Tesla T4 GPU, and Intel Xeon CPU. For dynamic power-accuracy tradeoff analysis, we utilize a run-time configurable approximate multiplier PAM [11], which supports up to 11 approximation levels for dynamic power-accuracy tradeoffs. To establish a baseline for comparison, we implement the approach from [6] and evaluate our XAIedge framework against this baseline. For evaluating the power consumption on TPU, we implemented the proposed approximate XAI hardware acceleration methods in Verilog HDL and synthesized them with Synopsys Design Compiler on UMC 40 nm library. The gate-level power evaluation is obtained using Synopsys PrimeTime based on the switching activity generated by ModelSim. For GPU, the power consumption of the proposed methods is estimated based on the power consumed by the approximate multiplier on $8 \times 8$ tiles per industry standard.

### B. Power Consumption of XAI Hardware Accelerators

We evaluated the power consumption of accurate and approximate XAI hardware acceleration on CPU, GPU, and TPU across 60 trials, as shown in Fig. 2. The notations xCPU, xGPU, and xTPU represent approximate XAI hardware acceleration on CPU, GPU, and TPU, respectively. We repeated the experiments for model distillation, Shapley analysis, and integrated gradients across 100 trials for each accelerator. To contextualize these results, we implemented the baseline framework for accurate XAI hardware acceleration from [6] and performed a comparative analysis. Our results show that in terms of energy efficiency, GPU-based approximate XAI hardware acceleration outperforms CPU-based execution, while TPU surpasses both GPU and CPU. Notably, power consumption is reduced by nearly 2× on each accelerator for XAI hardware acceleration after applying approximate computing. This improvement is attributed to the parallel computing capabilities of both GPU and TPU, which help balance workloads across cores. TPU delivers superior performance per watt, primarily because it uses 8-bit integer operations instead of 32-bit floating-point calculations. This significantly reduces the hardware size and hence, the power consumption of the TPU. While both GPU and TPU benefit from parallel computing, GPUs may not achieve the same level of specialization and efficiency as TPUs in tensor-based computations [16]. Our XAIedge framework consistently outperforms the baseline method in power efficiency across all hardware platforms, with the improvement being most pronounced on TPU, where our approach for XAI hardware achieves nearly **2×** lower power consumption due to optimized matrix operations and hardware-aware execution strategies.

TABLE I: Classification accuracy of different neural network models on CPU, GPU, and TPU-based hardware accelerators.

| Model | Model Distillation | | | Shapley Values | | | Integrated Gradients | | |
|---|---|---|---|---|---|---|---|---|---|
| | CPU | GPU | TPU | CPU | GPU | TPU | CPU | GPU | TPU |
| VGG16 | 94 | 92 | 96 | 94 | 90 | 95 | 88 | 97 | 93 |
| ResNet50 | 78 | 86 | 87 | 78 | 77 | 76 | 78 | 83 | 88 |
| MobileNetV2 | 96 | 93 | 97 | 91 | 97 | 94 | 95 | 90 | 98 |

TABLE II: Time Efficiency of outcome Interpretation through approx. model distillation, shapley values, and integrated gradients on CPU, GPU, and TPU accelerators. X/Y represents time efficiency on approx. vs. accurate accelerators.

| Model | Model Distillation | | | Shapley Values | | | Integrated Gradients | | |
|---|---|---|---|---|---|---|---|---|---|
| | CPU | GPU | TPU | CPU | GPU | TPU | CPU | GPU | TPU |
| VGG19 | 576.3/574.8 | 32.4/32.6 | 15.7/15.8 | 599/597 | 79.4/79.1 | 19.5/19.2 | 456.51/455.51 | 18.8/18.4 | 4.77/4.7 |
| ResNet50 | 1476.2/1474 | 182/182.3 | 38.9/38.6 | 56.8/54.6 | 14.12/14.13 | 3.9/3.6 | 21.5/19.3 | 1.8/1.82 | 1/1.1 |
| MobileNetV2 | 1846/1844 | 193/192 | 48/46.4 | 75.6/75.1 | 29/28.53 | 6.4/6.15 | 31.45/30.99 | 3.33/3.23 | 1.3/1.1 |

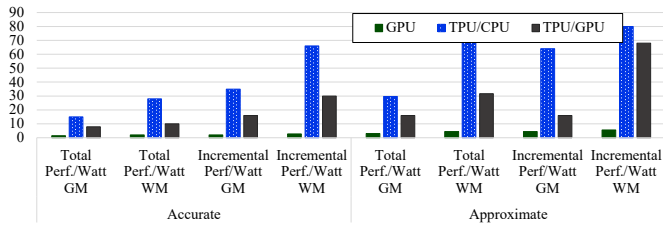

Fig. 3: Relative performance per watt of GPU (green bar) and TPU over CPU (blue bar), and TPU over GPU (gray bar) using model distillation.

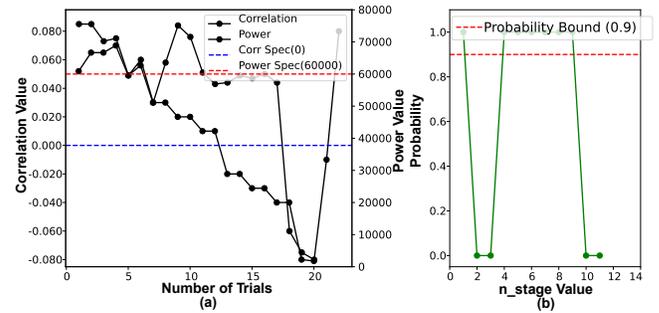

Fig. 4: Approximate integrated gradients with MILP solver on the TPU.

We further evaluated the geometric mean (GM) and weighted mean (WM) performance per watt for GPU and TPU relative to CPU. We calculated performance per watt in two different ways. The first method considers the total power consumed by the host CPU along with the actual execution performance per watt for the GPU or TPU. The second method, incremental performance per watt, excludes the host CPU power and reflects the actual power consumption of GPU or TPU during acceleration. Fig. 3 presents the performance per watt across these configurations. For accurate model distillation, the GPU implementation achieves total performance per watt $1.5\times$ GM and $2\times$ WM compared to the CPU. TPU achieves comparatively higher improvements, with total performance per watt $15\times$ GM and $28\times$ WM over CPU, and $7.9\times$ GM and $10\times$ WM over GPU. In terms of incremental performance per watt, TPU further outperforms CPU by $35\times$ GM and $66\times$ WM, and GPU by $16\times$ GM and $30\times$ WM. For approximate model distillation, performance per watt is reduced by $2\times$. For example, the GPU implementation achieves total performance per watt $0.77\times$ GM and $1.1\times$ WM compared to the CPU. TPU again outperforms both the CPU, with $7.4\times$ GM and $17.7\times$ WM, and the GPU, with $4\times$ GM and $7.9\times$ WM, in total performance per watt. For incremental performance per watt, TPU maintains its dominance in energy efficiency over the CPU with $16\times$ GM and $34\times$ WM, and over the GPU with $4\times$ GM and $17\times$ WM. These results confirm that our proposed framework not only aligns with the findings of [6] but also extends them by incorporating additional optimizations with approximate computing on hardware accelerators, particularly

### C. Time Efficiency of Outcome Interpretation

The average time required to perform outcome interpretation using three accurate and approximate XAI hardware acceleration methods for every 10 input-output pairs is presented in Table II. For approximate model distillation, TPU implementation significantly outperforms the others, achieving $36.6\times$ and $2.06\times$ faster execution than the CPU and GPU implementations, respectively, for VGG19. This performance gap widens for ResNet50, with speedups of $37.8\times$ over the CPU and $4.67\times$ over the GPU. Similar trends are observed across all XAI methods. Notably, the slightly higher average time observed in approximate model distillation is due to an additional 2-second overhead introduced by the lightweight optimization algorithm. Despite this, the rapid execution of outcome interpretation (within seconds) demonstrates the potential of XAIedge for integration into time-critical applications, marking a significant advancement in real-time XAI capabilities.

### D. Optimization of Approximation Levels

A key feature of the XAIedge framework is its ability to dynamically optimize the levels of approximation in XAI algorithm computations. This optimization is essential for balancing energy efficiency and output quality at each stage of the FFT computation, as outlined in Algorithm 4. The MINLP solver determines the optimal approximation level by considering the current input data, computation stage, and predefined PSNR and energy thresholds, enabling fine-grained control throughout the process. PSNR is commonly used in signal processing applications such as FFT to quantify the signal quality after approximation. The energy threshold ensures that the selected approximation levels yield meaningful energy savings. By dynamically optimizing approximation levels, XAIedge can adapt to varying input characteristics and computational stages, potentially enhancing overall performance compared to static methods. To validate the effectiveness of XAIedge, we conducted 100 trials of approximate Integrated Gradients using a diverse set of input images. We evaluated correlation factor, with the correlation bound $c_t$ to 0.0, $p_t$ to 6K, and $\mathcal{P}_t$ to 0.9. Fig. 4a illustrates the correlation factor and power consumption for the initial 25 trials, while Fig. 4b illustrates that approximation levels 3 to 10 consistently satisfy the $p_t$ constraint. Figs. 5a and 5b present interpretations of network intrusion using Shapley analysis and model distillation, with approximation levels 3 and 4, respectively, selected by our MINLP solver. These results underscore XAIedge's ability to maintain high accuracy while significantly reducing power consumption through intelligent approximation-level selection.

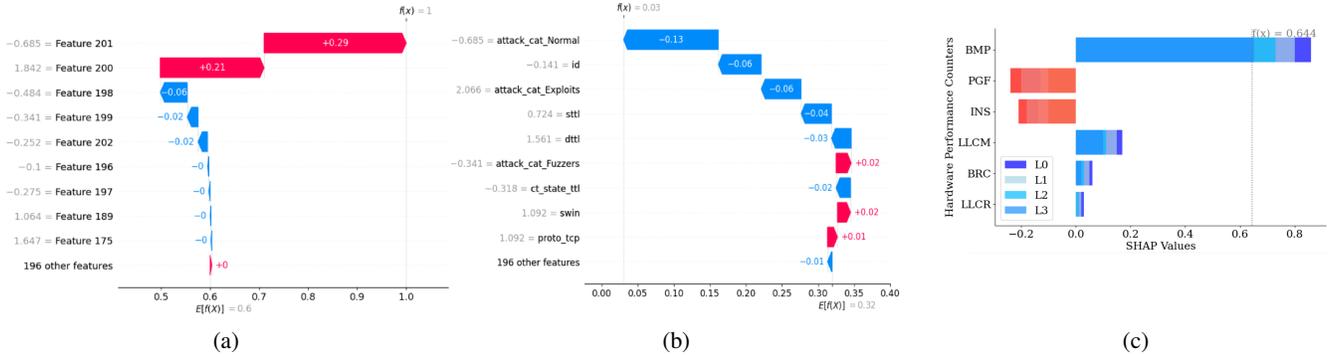

(a) (b) (c)

Fig. 5: (a) Interpretation of network intrusions using Shapley Analysis with approximation level 3 determined through MINLP solver, (b) Interpretation of network intrusions using Model Distillation with approximation level 4 determined through MINLP solver, (c) Shapley analysis for a Spectre attack: L0, L1 denote accurate and approximate analyses, respectively, with L2 (lower) & L3 (higher) indicating approximation levels.

|    | C1    | C2    | C3    | C4    | C5    |
|----|-------|-------|-------|-------|-------|
| R1 | 092c  | 31d   | 21a4  | d401  | bcd1  |
| R2 | 2244  | cc0f  | a6ab  | 7721  | 009a  |
| R3 | 0000  | 0004  | 00a1  | 0000  | 022c  |
| R4 | 00de  | bf31  | b0a1  | 2e03  | 1028  |
| R5 | 028a  | 0800  | 0028  | 0137  | 8c31  |
|    | 0.012 | 0.215 | 0.093 | 0.779 | 0.001 |
|    | 0.010 | 0.222 | 0.100 | 0.650 | 0.020 |

Fig. 6: Interpretation of MIRAI malware traced signals

*E. Outcome Interpretation Examples*

This section presents examples of outcome interpretation for our proposed XAI algorithms in different scenarios. Fig. 5c presents a waterfall plot for comparing accurate and approximate Shapley analysis methods with Spectre vulnerabilities, featuring hardware counters such as BMP and PGF. The plot shows the influence of each attribute on model predictions, with red color indicating positive contributions and blue color indicating negative ones. BMP and PGF emerge as significant predictors with induced page faults disrupting patterns, but the model compensates by prioritizing BMP. The significance of BMP remains higher than other attributes across varying approximation levels. However, approximate Shapley analysis at level L1 shows less significance compared to the accurate L0 result. Approximation levels from L2 (2-bit) to L3 (3-bit) maintain attribute significance despite approximation errors. These results demonstrate the effectiveness of our method under resource-constrained conditions.

Fig. 6 presents a trace table generated using the data from the MIRAI dataset. This table reveals the hexadecimal values in registers across specific clock cycles. The approximate model distillation method identifies the contribution of each cycle. Cycle C2 has the highest weight because "ATTACK VECTOR" variable is set at this cycle. This variable controls the bot's operational mode, such as initiating UDP or DNS attacks. Identifying this critical cycle improves detection confidence and helps isolate the malware source. Although the weights from approximate model distillation differ from those of the accurate model, their order of significance remains consistent. The analysis shows that despite precision differences between standard and approximate FFTs, the relative importance of each cycle is preserved. This observation highlights the role of the MINLP solver in selecting appropriate approximation levels.

Fig. 7 illustrates the outcome interpretation for integrated gradient, featuring the original image 7(a), the standard gradient map 7(b), and integrated gradient maps across various approximation levels. The pixel intensity indicates their significance in influencing the model's prediction. Our results show that the standard gradients inadequately capture the critical features due to their sensitivity to noise, leading to a dispersed distribution of attribution values. On the other hand, the integrated gradients provide a comprehensive view of the factors contributing to model decisions. Our proposed approximate integrated gradients method utilizes PAM [11], which is as accurate as a 32-bit floating-point multiplier at approximation level 11. Approximation level 11 demonstrates a good correlation with the expected attribution outcomes. Interestingly, the interpretation quality varies with the approximation levels: level 1 accurately captures critical features, level 2 suppresses them, level 4 highlights unwanted features, and level 6 achieves a good outcome interpretation. At level 1, PAM behaves similarly to a fixed-point approximate multiplier. Although higher levels in PAM are intended to increase PSNR, we observe that the PSNR does not significantly vary across approximation levels in this case. This indicates that correlation can serve as a more sensitive metric for evaluating the interpretability quality in approximate XAI. In the future, we plan to extend this work by further exploring the correlation between attribution values and ground truth feature importance at various approximation levels.

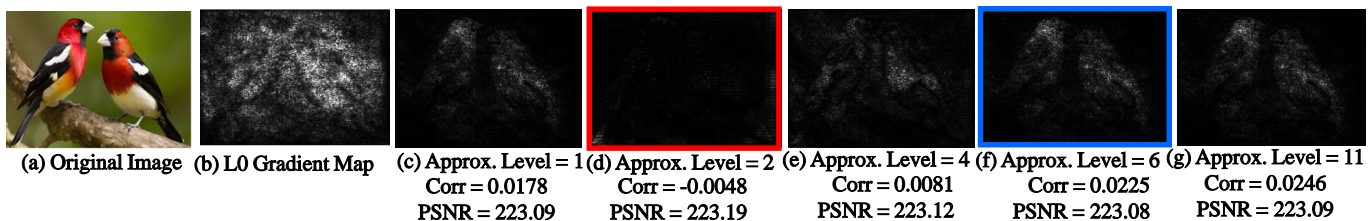

Fig. 7: Explainability of the original image (a), the gradient map (b), and integrated gradient maps with approximations levels (c) 1, (d) 2, (e) 4, (f) 6, (g) 11.

## V. Conclusion

XAI enhances transparency by framing interpretability as an optimization problem, which is a computationally intensive process that limits the feasibility of XAI for real-time applications. In this paper, we introduce XAIedge, a novel framework that leverages approximate computing in XAI algorithms for their energy-efficient hardware acceleration. XAIedge translates XAI algorithms into matrix computations and exploits Fourier transform and approximate computing for TPU-based edge devices. This facilitates real-time interpretations through parallel approximate computations. Our results show that XAIedge enhances energy efficiency by $2\times$ compared to existing accurate XAI hardware acceleration techniques.

## VI. Acknowledgments

This material is based upon work supported by the National Science Foundation (NSF) under Award Numbers: CCF-2323819. Any opinions, findings, conclusions, or recommendations expressed in this publication are those of the authors and do not necessarily reflect the views of the NSF.

## References

[1] B. Gulmezoglu, "XAI-based microarchitectural side-channel analysis for website fingerprinting attacks and defenses," *IEEE transactions on dependable and secure computing*, vol. 19, no. 6, pp. 4039–4051, 2021.

[2] Z. Pan, J. Sheldon, and P. Mishra, "Hardware-assisted malware detection and localization using explainable machine learning," *IEEE Transactions on Computers*, vol. 71, no. 12, pp. 3308–3321, 2022.

[3] T. Senevirathna, B. Siniarski, M. Liyanage, and S. Wang, "Deceiving post-hoc explainable ai (xai) methods in network intrusion detection," in *2024 IEEE 21st Consumer Communications Networking Conference (CCNC)*, 2024, pp. 107–112.

[4] R. Mitchell, E. Frank, and G. Holmes, "Gputreeshap: massively parallel exact calculation of shap scores for tree ensembles," *PeerJ Computer Science*, vol. 8, p. e880, 2022.

[5] H. Zhou, B. Zhang, R. Kannan, V. Prasanna, and C. Busart, "Model-architecture co-design for high performance temporal gnn inference on fpga," in *2022 IEEE International Parallel and Distributed Processing Symposium (IPDPS)*. IEEE, 2022, pp. 1108–1117.

[6] Z. Pan and P. Mishra, "Hardware acceleration of explainable machine learning," in *Proceedings of the 2022 Conference & Exhibition on Design, Automation & Test in Europe*, ser. DATE '22. Leuven, BEL: European Design and Automation Association, 2022, p. 1127–1130.

[7] A. Bhat, A. S. Assoa, and A. Raychowdhury, "Gradient backpropagation based feature attribution to enable explainable-ai on the edge," in *2022 IFIP/IEEE 30th International Conference on Very Large Scale Integration (VLSI-SoC)*. IEEE, 2022, pp. 1–6.

[8] A. Bhat and A. Raychowdhury, "Non-uniform interpolation in integrated gradients for low-latency explainable-ai," in *2023 IEEE International Symposium on Circuits and Systems (ISCAS)*. IEEE, 2023, pp. 1–5.

[9] K. Alex, "Learning multiple layers of features from tiny images," *https://www.cs.toronto.edu/kriz/learning-features-2009-TR.pdf*, 2009.

[10] K. Angrishi, "Turning internet of things (iot) into internet of vulnerabilities (iov): Iot botnets," *arXiv preprint arXiv:1702.03681*, 2017.

[11] C. Chen, W. Qian, M. Imani, X. Yin, and C. Zhuo, "Pam: A piecewise-linearly-approximated floating-point multiplier with unbiasedness and configurability," *IEEE Transactions on Computers*, vol. 71, no. 10, pp. 2473–2486, 2021.

[12] T. Yu, B. Wu, K. Chen, C. Yan, and W. Liu, "Toward efficient retraining: A large-scale approximate neural network framework with cross-layer optimization," *IEEE Transactions on Very Large Scale Integration (VLSI) Systems*, 2024.

[13] A. Siddique and K. A. Hoque, "Is approximation universally defensive against adversarial attacks in deep neural networks?" in *2022 Design, Automation & Test in Europe Conference & Exhibition (DATE)*. IEEE, 2022, pp. 364–369.

[14] ——, "Exposing reliability degradation and mitigation in approximate dnns under permanent faults," *IEEE Transactions on Very Large Scale Integration (VLSI) Systems*, vol. 31, no. 4, pp. 555–566, 2023.

[15] J. Kronqvist, D. E. Bernal, A. Lundell, and I. E. Grossmann, "A review and comparison of solvers for convex minlp," *Optimization and Engineering*, vol. 20, pp. 397–455, 2019.

[16] J. Mendez, K. Bierzynski, M. P. Cuéllar, and D. P. Morales, "Edge intelligence: concepts, architectures, applications, and future directions," *ACM Transactions on Embedded Computing Systems (TECS)*, vol. 21, no. 5, pp. 1–41, 2022.